\documentclass[sigconf]{acmart}
\usepackage{natbib}
\usepackage{booktabs} 
\usepackage{breqn}
\setcopyright{rightsretained}
\newcommand{\Lagr}{\mathcal{L}}
\newcommand{\E}{\mathbb{E}}
\def\acmSubmissionID{122}
\acmDOI{10.475/123_4}

\acmISBN{123-4567-24-567/08/06}

\acmConference[ICVGIP'22]{13th Indian Conference on Computer Vision, Graphics and Image Processing}{December 2022}{Gandhinagar, India}
\acmYear{2022}
\copyrightyear{2022}

\acmPrice{15.00}

\begin{document}
\title{Variational Augmentation for Enhancing Historical Document Image Binarization}
\titlenote{Produces the permission block, and
  copyright information}

\author{Avirup Dey}
\affiliation{%
  \institution{Jadavpur University}
  \city{Kolkata}
  \state{West Bengal}
  \country{India}
  \postcode{700032}
}
\author{Nibaran Das}
\affiliation{%
  \institution{Jadavpur University}
  \city{Kolkata}
  \state{West Bengal}
  \country{India}
  \postcode{700032}
}
\author{Mita Nasipuri}
\affiliation{%
  \institution{Jadavpur University}
  \city{Kolkata}
  \state{West Bengal}
  \country{India}
  \postcode{700032}
}
\renewcommand{\shortauthors}{Dey et. al.}

\begin{abstract}
Historical Document Image Binarization is a well-known seg-
mentation problem in image processing. Despite ubiquity, traditional
thresholding algorithms achieved limited success on severely degraded
document images. With the advent of deep learning, several segmentation models were proposed that made significant progress in the field but were limited by the unavailability of large training datasets. To mitigate this problem, we have proposed a novel two-stage framework- the first of which comprises a generator that generates degraded samples using variational inference and the second being a CNN-based binarization network that trains on the generated data. We evaluated our framework on a range of DIBCO datasets, where it achieved competitive results against previous state-of-the-art methods.
\end{abstract}

%
%



\maketitle

\section{Introduction}
One of the most important preprocessing steps for the analysis of document images is their binarization. The classification of image pixels into text and non-text regions greatly aids systems that reside on the other end of the pipeline, generally optical character recognition algorithms. While binarization is often performed with the use of a simple threshold that dictates how the image pixel would be mapped in the output image, advanced binarization algorithms are needed when the task involves the restoration of document images in conjunction with binarization to obtain an ideal text/non-text demarcation.
\par
Over the years, many restoration algorithms have been proposed focussing on the removal of degradations found in document images. Traditional Image processing-based algorithms like Otsu\cite{otsu1979threshold}, Niblack \cite{niblack1985introduction} and Sauvola\cite{sauvola2000adaptive} deal with minor degradations successfully and are particularly popular in this domain. These methods, however, do not perform successful binarization when the degradations are severe, typically those observed in historical document images. Binarization of severely degraded document images has been addressed recently with the help of deep learning algorithms, especially CNNs \cite{lecun2015deep}, that have shown significant promise in this domain. Tenseymer and Martinez proposed a fully connected network to binarize document images in \cite{tensmeyer2017document}, treating binarization as a special case of semantic segmentation. Calvo-Zaragoza and  A.J. Gallego proposed a deep encoder-decoder architecture, outperforming previous binarization algorithms in \cite{calvo2019selectional}. Vo et al.\cite{vo2018binarization} proposed a hierarchical deep learning framework that achieved state-of-the-art results on multiple datasets. Later He and Schomaker\cite{he2019deepotsu} proposed an iterative CNN framework and achieved similar results. Suh \textit{et al.} \cite{suh2020two} proposed a two-stage adversarial framework for binarizing RGB images. The first stage trains an adversarial network for each channel of the RGB image to extract foreground information from small local image patches by removing background information for document image enhancement. The second stage learns the local and global features to produce the final binarized output. This algorithm outperformed previous methods and achieved state-of-the-art performance on multiple datasets.
\par
Although the previous methods can perform successful binarizations of degraded documents, they are constrained to grayscale inputs and require voluminous training data. To deal with the lack of adequate training data, Bhunia \textit{et al.}\cite{bhunia2019improving} introduced a texture augmentation network (TANet) in conjunction with a binarization network that generates degraded document images with the help of style transfer. This method achieves competitive performance with previous methods but is constrained by the one-to-one mapping of style transfer-based approaches. Capobianco and Marinai \cite{capobianco2017docemul} proposed a toolkit to generate synthetic document images by combining text and degradations from a given palette. One could iteratively generate multiple images from a given text but this requires a lot of manual cherry-picking of degradation styles. 
\par
In this paper we propose a novel two-stage framework addressing both the data scarcity in this domain and the binarization of severely degraded document images containing multiple artefacts in the form of stains and bleed-through. In the first stage, we employ an augmentation network, Aug-Net, inspired by the BicycleGAN \cite{zhu2017toward}, that generates novel degraded samples from a single image-ground truth pair using variational inference. Combining the strengths from GANs \cite{goodfellow2014generative} and VAEs \cite{kingma2013auto}, this network encodes the content and style of the input image into a probability distribution, allowing the model to generate multiple outputs from a single input by sampling from the distribution during inference. Additionally, it resolves the blurring effect of VAE by encouraging bijective similarity, making the generated samples more realistic.
\par
In the following stage, we employ a paired image to image translation network, Bi-Net, inspired by Pix2Pix \cite{isola2017image}, to binarize the images. While previous methods performed binarization from grayscale images, we work with RGB images, attempting to capture the nature of the degradation from multiple channels. Our architecture further employs PixelShuffle \cite{shi2016real} upsampling, replacing the traditional transpose convolutional layer. PixelShuffle uses Efficient Sub-Pixel Convolution (ESPC) that is computationally less expensive and has shown better results in upscaling images and videos.
\par
In our experiments, we trained our network on DIBCO 2009 \cite{gatos2009icdar}, DIBCO 2010 \cite{pratikakis2010h}, DIBCO 2011 \cite{6065249}, DIBCO 2013 \cite{pratikakis2013icdar}, and DIBCO 2017 \cite{8270159} datasets and tested it on DIBCO 2014 \cite{ntirogiannis2014icfhr2014}, DIBCO 2016 \cite{pratikakis2016icfhr2016}, DIBCO 2018\cite{inproceedings2018} and DIBCO 2019 \cite{inproceedings2019} using metrics like F-measure, pseudo F-measure, PSNR and DRD for evaluation of performance.
Our method achieves competitive performance on the said datasets against several state-of-the-art models, including Mesquita \textit{et. al} \cite{mesquita2015parameter} (winner of DIBCO 2014), Kligger and Tal \cite{kligler2018document} (winner of DIBCO 2016), Wei \textit{et. al.} \cite{xiong2018historical} (winner of DIBCO 2018) and Sarkar \textit{et. al.} \cite{bera2021non} (winner of DIBCO 2019).
\par
Our contributions can be summarised as follows:

\begin{itemize}
    \item We have used variational inference to generate novel images for training that can scale up the training data by many folds, thus eliminating the problem of limited training samples.
    
    \item We have used PixelShuffle upsampling instead of transpose convolutions and short skip connections in the generator of our binarization model in addition to the long symmetric skip connections. These modifications not only yielded better results but also helped in faster training.
    
    \item We have evaluated our model on multiple metrics against several state-of-the-art binarization algorithms, including the winners of previous DIBCO contests showing that our proposed method achieves significant improvements over existing algorithms.
    
\end{itemize}

The paper is organised as follows: Section 2 summarizes the relevant literature, Section 3 covers our proposed methodology comprising the network architectures and training details, Section 4 and Section 5 cover our experimental results and ablation studies, respectively, and Section 6 concludes the paper.
\section{Related Works}
The overall goal of document binarization is to convert an input image into a two-tone version, enabling easy demarcation of text and non-text regions to enhance the information that can be extracted from the same. Towards this goal, a plethora of methods have been proposed, broadly classified into two groups, classical image processing-based methods and deep learning-based algorithms.

\subsection{Image Processing Based Methods}
Traditional image processing based methods like those proposed in \cite{otsu1979threshold}, \cite{niblack1985introduction}, and \cite{sauvola2000adaptive} formed the baseline of research in the domain of document image binarization. One of the most popular algorithms was proposed by Otsu \textit{et al.} in \cite{otsu1979threshold}, which computes a threshold that minimizes the intra-class variance and maximizes the inter-class variance of two pre-assumed classes. It selects the global threshold based on a histogram, and owing to its simplicity this algorithm is very fast. However, it is sensitive to deep stains, non-uniform backgrounds and bleed-through degradations. 
\par
 In order to solve this problem, local adaptive threshold methods were proposed, such as Sauvola\cite{sauvola2000adaptive}, Niblack\cite{niblack1985introduction} and AdOtsu\cite{moghaddam2012adotsu}. These methods compute the local threshold for each pixel based on the mean and standard deviation of a local area around the pixel. Following these methods, several binarization methods have been proposed. Gatos \textit{et al.} \cite{gatos2006adaptive} used a Wiener filter to estimate the background and foreground regions in a method that employs several post-processing steps to remove noise in the background and improve the quality of foreground regions. Su \textit{et al.} \cite{su2012robust} introduced a map combining the local image contrast and the local image gradient; the local threshold is estimated based on the values on the detected edges in a local region. Pai \textit{et al.} \cite{pai2010adaptive} proposed an adaptive window-size selection method based on the foreground characteristics. Jia \textit{et al.} \cite{jia2018degraded} employed the structural symmetry of strokes to compute the local threshold. Xiong \textit{et. al.} \cite{xiong2021enhanced} proposed an entropy-based formulation to segregate text from the image background and employed energy-based segmentation to binarize images. They achieved competitive results on DIBCO benchmarks.
 \par
 However, these adaptive binarization methods require many empirical parameters that need to be manually fine-tuned and are still not satisfactory for use with highly degraded and poor-quality document images.

\subsection{Deep Learning Based Methods}
In recent years, convolutional neural networks (CNNs) \cite{lecun2015deep} have achieved several milestones in a variety of tasks in computer vision. Tenseymer and Martinez \cite{tensmeyer2017document} used a fully convolutional network for document image binarization at multiple image scales. 
A deep auto-encoder–decoder architecture model was proposed by Calvo-Zaragoza and  A.J. Gallego. Vo \textit{et al.} \cite{calvo2019selectional} proposed a hierarchical deep supervised network to predict the full foreground map through the results of multi-scale networks; this method achieved state-of-the-art performance on several benchmarks. 
He and Schomaker \cite{he2019deepotsu} introduced an iterative CNN-based framework and achieved performance similar to that of \cite{vo2018binarization}'s method. Soibgui \textit{et al.} \cite{souibgui2022docentr} developed a transformer \cite{vaswani2017attention} based encoder-decoder framework which achieved results comparable to the previous method.
\par
Recently, generative adversarial networks (GANs) \cite{goodfellow2014generative} have emerged as a class of CNN models approximating real-world images, achieving significant performance image synthesis. Over the years, several variants have been proposed for image translation tasks as well. In GANs, a generator network competes against a discriminator network that distinguishes between generated and real images. 
Unlike the original GAN, cGAN \cite{mirza2014conditional} trains the generator not only to fool the discriminator but also to condition it on additional inputs, such as class labels, partial data, or input images. Based on this principle, Isola \textit{et al.} \cite{isola2017image} proposed Pix2Pix GAN for the general purpose of image-to-image translation. Bhunia \textit{et al.} \cite{bhunia2019improving} proposed a texture augmentation network to augment the training datasets and handle image binarization using a cGAN structure. 
Zhao \textit{et al.} \cite{zhao2019document} proposed a cascaded network based on Pix2Pix GAN to combine global and local information. Suh \textit{et al.} \cite{suh2020two} proposed a two-stage GAN framework, employing adversarial networks to remove noise from each channel separately in the first stage and fine-tune the output using a similar network in the second stage.

\section{Proposed Methodology}
\subsection{Overview}
The application of deep learning algorithms in the field of historical document image binarization is limited due to the dearth of training data. In the past, researchers had to either curate and preprocess images manually from multiple sources as shown in \cite{capobianco2017docemul} or use style transfer-based augmentation as shown by Bhunia \textit{et. al.} \cite{bhunia2019improving} and Kumar \textit{et. al.} \cite{kumar2021udbnet}. Despite their merits, both methods suffer from the lack of diversity of degradations real historical documents might have. \\

\begin{figure}
    \centering
    \includegraphics[width=\linewidth]{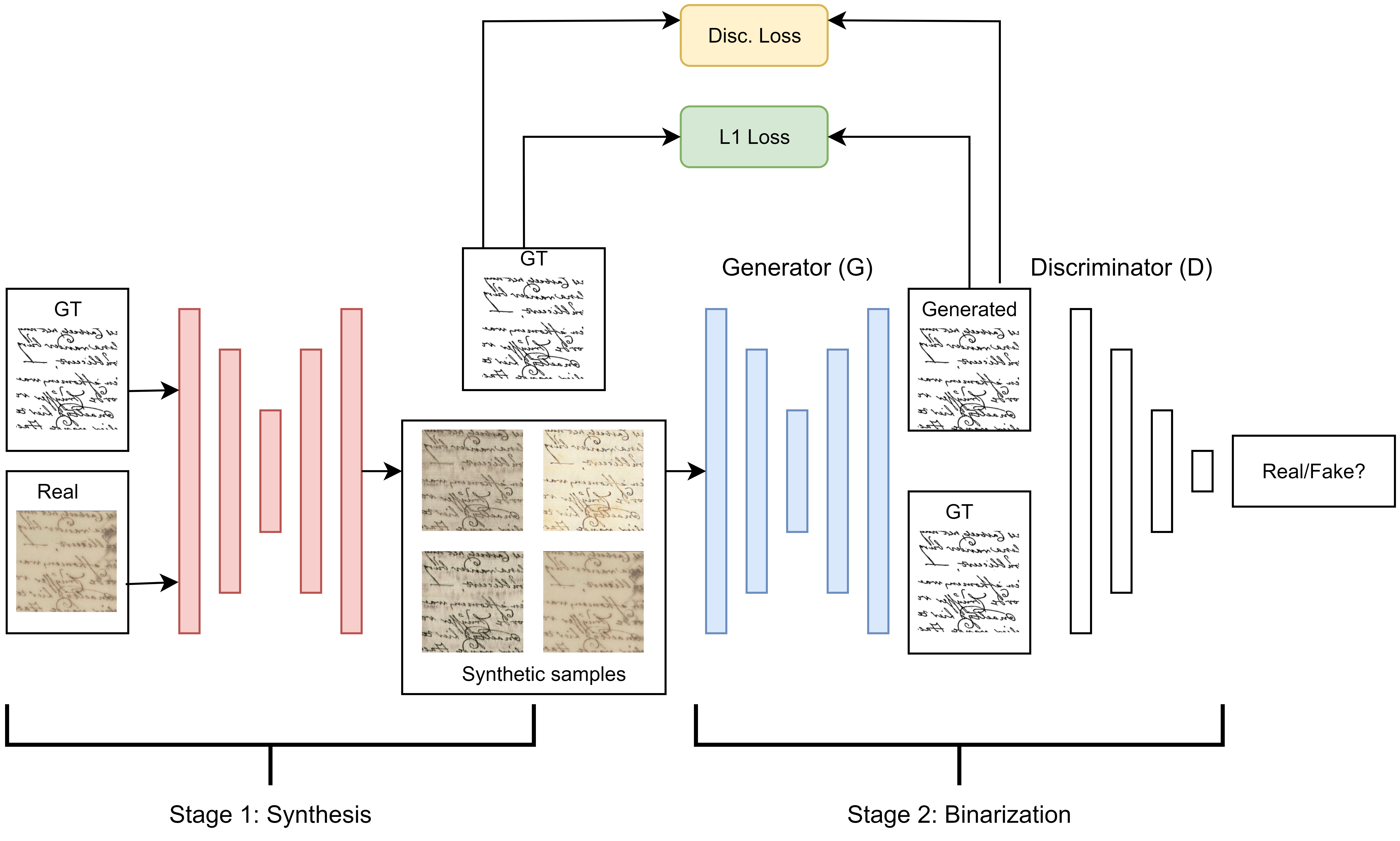}
    \caption{Training the Binarization Network: Stage-I comprises the variational generator of the augmentation network (Aug-Net) that synthesizes novel training samples from a given input. Stage-II is comprises the binarization network (Bi-Net) that is trained to transform the degraded images into binary masks. }
    \label{fig:block}
\end{figure}

The strength of our method lies in the generation of novel synthetic data in situ while training the binarization model as shown in Fig.\ref{fig:block}.
In the first stage, an augmentation network, inspired by Bicycle-GAN \cite{zhu2017toward}, generates multiple degraded samples from an input-ground truth pair via variational inference. These synthetic samples are, in turn, used to train our binarization network in the next stage.
We have employed a paired image translation procedure, following the work of Isola \textit{et. al.} \cite{isola2017image}.

\subsection{Network Architectures}
\subsubsection{(A) Augmentation Network (Aug-Net)}
\begin{figure}
    \centering
    \includegraphics[width=\linewidth]{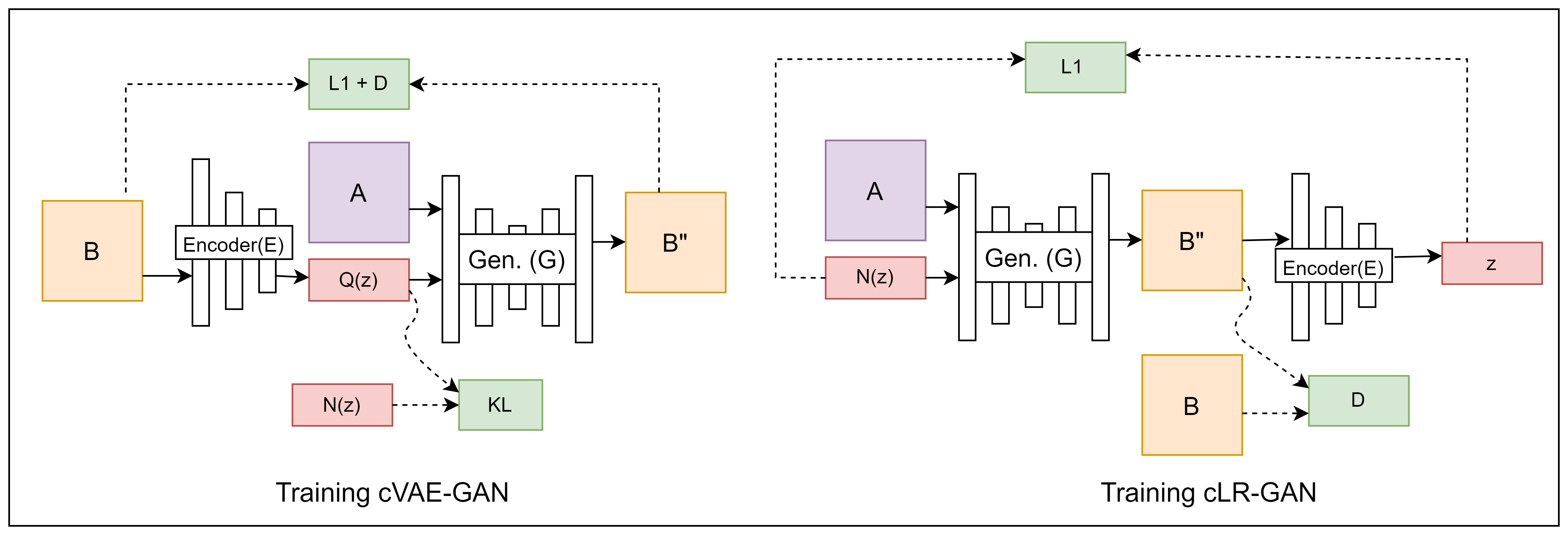}
    \caption{Training the Augmentation Network: cVAE-GAN starts from a ground truth target image B and encodes it into the latent space. The generator then attempts to map the input image A along with a sampled z back into the original image B. cLR-GAN randomly samples a latent code from a known distribution, uses it to map A into the output B, and then tries to reconstruct the latent code from the output.}
    \label{fig:bicyclegan}
\end{figure}

~\\
The network comprises two components - cVAE-GAN \cite{larsen2016autoencoding} and cLR-GAN \cite{dumoulin2016adversarially} which complement each other in the training process. cVAE-GAN learns the encoding from real data, but a random latent code may not yield realistic images at test time, and the KL loss may not be well optimized. More importantly, the discriminator does not have a chance to see results sampled from the prior during training. On the other hand, in cLR-GAN, the latent space is easily sampled from a simple distribution, but the generator is trained without the benefit of observing the ground truth input-output pairs. Thus, combining these helps us produce results that are diverse as well as realistic.
Fig.\ref{fig:bicyclegan} outlines the training procedure of our augmentation network.

We use the trained generator of the cVAE-GAN to generate new degraded images while training our binarization network.
\newline

\subsubsection{(B) Binarization Network (Bi-Net)}
\begin{figure}
    \centering
    \includegraphics[width=\linewidth]{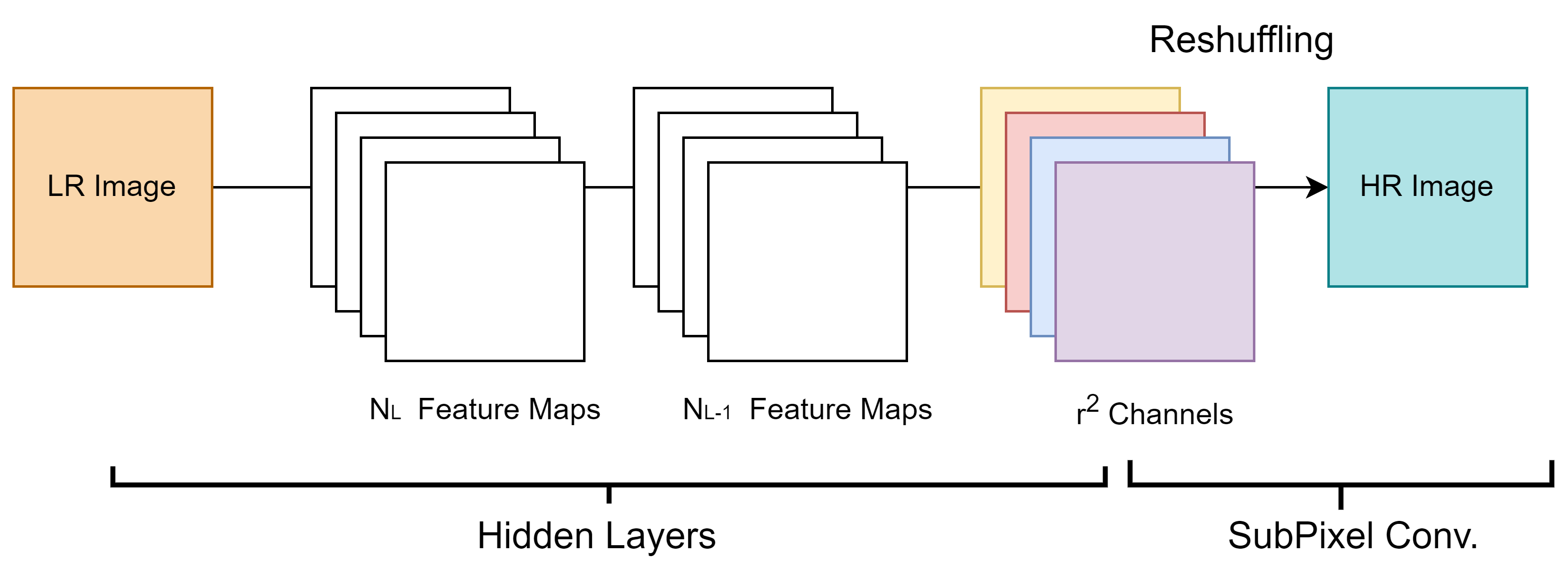}
    \caption{Efficient Sub-Pixel Convolution: A low resolution image is passed though convolution layers to generate $r^{2}$ feature maps which are aggregated to give a higher resolution output. }
    \label{fig:subconv}
\end{figure}

~\\
Our binarization network is inspired by the well-known Pix2Pix GAN \cite{isola2017image} which was first proposed by Isola \textit{et. al.} with some changes in the generator architecture. Firstly, we have built the U-Net generator on top of a ResNet \cite{he2016deep} backbone that helps in stabilizing the training. Secondly, we have swapped the dilated convolutions in the upsampling blocks with PixelShuffle \cite{shi2016real} that produce sharper outputs.\\
The PixelShuffle layer implements Efficient Sub-Pixel Convolution. The operation is described by the following equation:
\begin{equation}
I^{HR}\;=\;f^{L}(I^{LR})\;=\;PS(W_{L}\;*\;f^{L-1}(I^{LR})\;+\;b_{L})
\end{equation}
where $I^{HR}$ denotes the high resolution image, $I^{LR}$ denotes the low resolution image, $f^{L}$ is the convolution kernel of the L-th layer and PS is an periodic shuffling operator that rearranges the elements of a $H \times W \times C \cdot r^{2}$  tensor to a tensor of shape $rH \times rW \times C$. This has been illustrated in Figure \ref{fig:subconv}.\\

We employed a PatchGAN \cite{li2016precomputed} discriminator with an output scale of $70 \times 70$.

\subsection{Objective Functions}
\subsubsection{(A) For Augmentation Network}~\\
We train our augmentation network on a combination of losses that includes KL Divergence, adversarial loss and a regularizer $L_{1}$ loss as proposed by \cite{zhu2017toward}. The objective is given by:
\begin{multline*}
    G^{*}, E^{*}\;=\;arg\;min_{G,E}\;max_{D}\:\Lagr_{GAN}^{VAE}(G,D,E)\;+\;\lambda\Lagr^{VAE}_{1}(G,E)\\
    \;+\;\Lagr_{GAN}(G,D)\;+\;\lambda_{latent}\Lagr_{1}^{latent}(G,E)\;+\;\lambda_{KL}\Lagr_{KL}(E)
\end{multline*}

$\lambda=10$, $\lambda_{latent}=0.5$ and $\lambda_{KL}=0.01$ were the optimum hyper-parameters. The length of the latent vector was taken as $|z|=8$.
\newline

\subsubsection{(B) For Binarization Network}~\\
We train our binarization network on a combination of the conditional adversarial loss as proposed by Isola \textit{et. al.} \cite{isola2017image}. and the $L_{1}$ loss. 
The conditional adversarial loss is given by: \\
\begin{equation}
\Lagr_{cGAN}\;=\;\E_{x, y}[log D(x, y)]\;-\;\E_{x, z}[log(1 - D(x, G(x, z)))]
\end{equation}
Conditioning the discriminator term with the input, $x$ forces the generator to produce images that are perceptually similar in structure. It also reduces blurring artefacts.
Adding an $L_{1}$ loss encourages the discriminator to force pixel-level accuracy in the generated images.\\
The overall objective can be written as:
\begin{equation}
G^{*}\;=\;arg\;min_{G}\;max_{D}\:\Lagr_{cGAN}(G, D)\;+\;\lambda\Lagr_{L1}(G)
\end{equation}
$\lambda$ is a hyper-parameter that can be tuned to weight the loss terms. We have trained our model with $\lambda\;=\;100$.

\subsection{Training}
Both the networks were trained with patches extracted from DIBCO 2009 \cite{gatos2009icdar}, 2010 \cite{pratikakis2010h}, 2011 \cite{6065249}, 2013 \cite{pratikakis2013icdar} and 2017 \cite{8270159} datasets. \\
The augmentation network was trained on $256\times256$ patches extracted from the said datasets for 6 epochs using the Adam optimizer.\\
For training the binarization network, patches of size $512\times512$ were extracted from each image to obtain 13320 samples. Each sample was further reduced to $256\times256$ using augmentations like random crop and resize.
The generator was pre-trained with just the $L_{1}$ loss for 5 epochs. The model was then trained adversarially for 20 epochs using the Adam optimizer ($\beta_{1}=0.5$ and $\beta_{2}=0.999$) with a learning rate of $2e^{-4}$.

\section{Experimental Results}

\begin{figure}
    \centering
    \fbox{\includegraphics[width=0.20\linewidth]{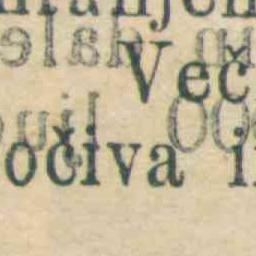}}\hspace{0.1cm}
    \fbox{\includegraphics[width=0.20\linewidth]{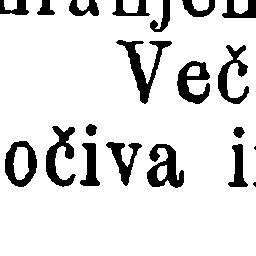}}\hspace{0.1cm}
    \fbox{\includegraphics[width=0.20\linewidth]{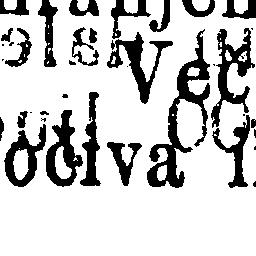}}\hspace{0.1cm}
    \fbox{\includegraphics[width=0.20\linewidth]{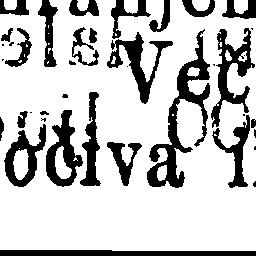}}\vspace{0.1cm}
    \fbox{\includegraphics[width=0.20\linewidth]{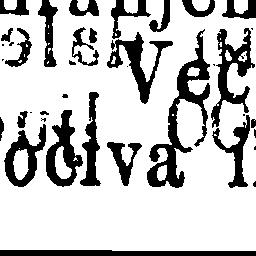}}\hspace{0.1cm}
    \fbox{\includegraphics[width=0.20\linewidth]{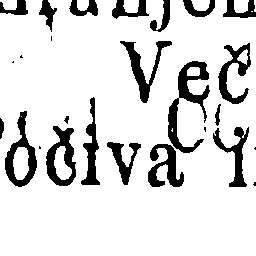}}\hspace{0.1cm}
    \fbox{\includegraphics[width=0.20\linewidth]{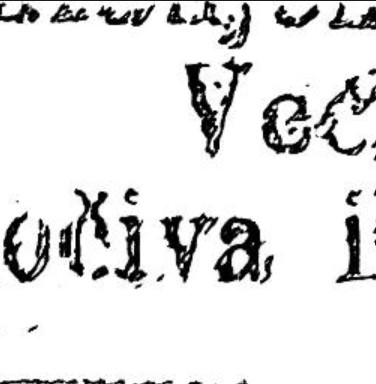}}\hspace{0.1cm}
    \fbox{\includegraphics[width=0.20\linewidth]{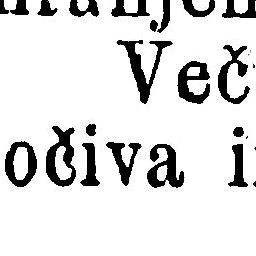}}
    \caption{From left to right, top to bottom: Input, Ground Truth, Otsu, Sauvola, Niblack, Suh, Bhunia, Ours}
    \label{fig:results}
\end{figure}

\subsection{Datasets}
All comparisons were carried out using DIBCO datasets. For evaluation purposes, we have used DIBCO 2014 \cite{ntirogiannis2014icfhr2014}, DIBCO 2016 \cite{pratikakis2016icfhr2016}, DIBCO 2018 \cite{inproceedings2018} and DIBCO 2019 \cite{inproceedings2019} datasets. \\
\newline
\textbf{Data Preparation:} Each DIBCO dataset has 10 degraded historical document images for evaluation. We extracted 5 random patches of dimension $256 \times 256$ from each image, giving us a total of 50 samples from each dataset for evaluation.\\
\newline
\subsection{Evaluation Metrics}
\begin{itemize}
    \item \textbf{F Measure:}\\
    \begin{equation}FM\;=\;\frac{2 \times Precision \times Recall}{Precision + Recall},\end{equation}
        where $Precision = \frac{TP}{TP+FN}$ and $Recall = \frac{TP}{TP+FP}$. TP, FP and FN denote true positive, false positive and false negative respectively.\\
    \item \textbf{Pseudo-F-Measure:}\\
    \begin{equation}pFM\;=\;\frac{2 \times Precision \times pRecall}{Precision + pRecall},\end{equation}
        where $pRecall$ denotes the fraction of the skeletonized ground truth image. \cite{pratikakis2010h}. \\
    \item \textbf{PSNR:}\\
     \begin{equation}PSNR\;=\;10\;log(\frac{C^{2}}{MSE}),\end{equation}
        where $MSE$ is the mean squared error between the two images. $C$ is the maximum pixel value. The images were normalized so here, $C=1.$\\ PSNR is a widely used metric to measure the similarity between two images.\\
    \item \textbf{DRD:}\\
    \begin{equation}DRD\;=\;\frac{\Sigma_{k}\; DRD_{k}}{NUBN}\end{equation}\\
    where $DRD_{k}$ is the distortion of the k-th flipped pixel, and $NUBN$ is the number of non-uniform $8 \times 8$ blocks in the ground truth image. \cite{ntirogiannis2014icfhr2014}.
            
\end{itemize}
\subsection{Results and Discussion}
From Table \ref{tab:maintable} we can see that our proposed model achieves competitive scores against previous state-of-the-art methods, and in some cases even surpasses them.\\
We also observe that, unlike the winning models of the DIBCO contests (\cite{kligler2018document}, \cite{xiong2018historical}, \cite{bera2021non}), our model not only maximises pixel-level accuracy (measured by F-Measure and PSNR) but also preserves perceptual quality of the binarized image (measured by DRD). This is a clear consequence of using a conditional GAN loss.\\
We compared our model to a range of traditional thresholding algorithms like Otsu \cite{otsu1979threshold}, Niblack \cite{niblack1985introduction} and Sauvola \cite{sauvola2000adaptive} as well as recent state-of-the-art like Vo \cite{vo2018binarization}, Xiong \cite{xiong2021enhanced} and Suh \cite{suh2020two}. The winning models of DIBCO 14 \cite{ntirogiannis2014icfhr2014}, DIBCO 16 \cite{pratikakis2016icfhr2016}, DIBCO 18 \cite{inproceedings2018} and DIBCO 19 \cite{inproceedings2019} have also been included in our experiments.\\
Bhunia's method \cite{bhunia2019improving}, Suh's method \cite{suh2020two}, Soibgui's method \cite{souibgui2020gan} and Cycle-GAN \cite{zhu2017unpaired} were trained on the datatsets we used for training our own model (DIBCO 2009 \cite{gatos2009icdar}, DIBCO 2010 \cite{pratikakis2010h}, DIBCO 2011 \cite{6065249}, DIBCO 2013 \cite{pratikakis2013icdar} and DIBCO 2017 \cite{8270159}) with the hyper-parameters set by their respective authors.

Bhunia \textit{et. al.'s} model \cite{bhunia2019improving}, Suh \textit{et. al.'s} model \cite{suh2020two}, Vo \textit{et.al.'s} model \cite{vo2018binarization} and He's \cite{he2019deepotsu} model were originally trained on PHIBD \cite{nafchi2013efficient} dataset that has over 100 images in addition to multiple DIBCO datasets. On the other hand, our training dataset consisted of only 74 DIBCO images. In effect, we trained our model on almost half the number of training images used by previous deep learning-based methods Hence, it demonstrates the strength of variational augmentation while training the binarization model.

\begin{table}
\caption{\label{tab:maintable}Comparative results on DIBCO datasets.}
\setlength{\tabcolsep}{3pt}
\centering
\begin{tabular}{|l|c|c|c|c|}
\hline
\multicolumn{1}{|c|}{\textbf{Method}} & \multicolumn{4}{c|}{\textbf{Evaluation Metrics}}                                                                                                            \\ \hline
\textbf{}                             & \multicolumn{1}{l|}{\textbf{F-Measure $\uparrow$}} & \multicolumn{1}{l|}{\textbf{pF-Measure $\uparrow$}} & \multicolumn{1}{l|}{\textbf{PSNR $\uparrow$}} & \multicolumn{1}{l|}{\textbf{DRD $ \downarrow$}} \\ \cline{2-5} 
                                      & \multicolumn{4}{c|}{\textbf{DIBCO 2014}}
                                       \\ \cline{2-5}
\textbf{Rank 1 \cite{mesquita2015parameter}}                       & \textbf{96.880}                         & \textbf{97.650}                          & \textbf{22.660}                    & \textbf{0.902}                    \\
\textbf{Otsu \cite{otsu1979threshold}}                         & 91.780                                  & 95.740                                   & 18.720                             & 2.647                             \\
\textbf{Niblack \cite{niblack1985introduction}}                      & 86.010                                  & 88.040                                   & 16.540                             & 8.260                             \\
\textbf{Sauvola \cite{sauvola2000adaptive}}                      & 86.830                                  & 91.800                                   & 17.630                             & 4.896                             \\
\textbf{Vo \cite{vo2018binarization}}                           & 95.970                                  & 97.420                                   & 21.490                             & 1.090                             \\
\textbf{He \cite{he2019deepotsu}}                           & 95.950                                  & 98.760                                   & 21.600                             & 1.120                             \\
\textbf{Xiong \cite{xiong2021enhanced}}                        & 96.770                                  & 97.730                                   & 22.470                             & 0.950                             \\
\textbf{Bhunia \cite{bhunia2019improving}}                       & 73.753                                  & 74.000                                   & 12.679                             & 7.850                             \\
\textbf{Suh \cite{suh2020two}}                          & 96.360                                  & 98.870                                   & 21.960                             & 1.070                             \\
\textbf{Soibgui \cite{souibgui2020gan}}                      & 96.020                                        & 95.300                                         & 19.870                                   & 4.100                                  \\
\textbf{Cycle-GAN \cite{zhu2017unpaired}}                    & 84.530                                  & 86.792                                   & 15.278                             & 5.993                             \\
\textbf{Ours}                         & 96.860                                  & 97.450                                   & 22.250                             & 1.530                             \\ \cline{2-5} 
\textbf{}                             & \multicolumn{4}{c|}{\textbf{DIBCO 2016}}                                                                                                                    \\ \cline{2-5} 
\textbf{Rank 1 \cite{kligler2018document}}                       & 87.610                                  & 91.280                                   & 18.110                             & 5.210                             \\
\textbf{Otsu \cite{otsu1979threshold}}                         & 85.660                                  & 88.860                                   & 16.260                             & 5.580                             \\
\textbf{Niblack \cite{niblack1985introduction}}                      & 72.570                                  & 73.510                                   & 13.260                             & 24.650                            \\
\textbf{Sauvola \cite{sauvola2000adaptive}}                      & 84.270                                  & 89.100                                   & 17.150                             & 6.090                             \\
\textbf{Vo \cite{vo2018binarization}}                           & 90.010                                  & 93.440                                   & 18.740                             & 3.910                             \\
\textbf{He \cite{he2019deepotsu}}                           & 91.190                                  & 95.740                                   & 19.510                             & 3.020                             \\
\textbf{Xiong \cite{xiong2021enhanced}}                        & 89.640                                  & 93.560                                   & 18.690                             & 4.030                             \\
\textbf{Das \cite{das2019statistical}}                        & 88.930                                  & 91.750                                   & 18.080                             & 4.120                             \\
\textbf{Bhunia \cite{bhunia2019improving}}                       & 65.525                                  & 65.145                                   & 12.595                             & 8.270                             \\
\textbf{Suh \cite{suh2020two}}                          & 92.240                                  & 95.950                                   & 19.930                             & 2.770                             \\
\textbf{Soibgui \cite{souibgui2020gan}}                      & 88.760                                        & 87.230                                         & 19.450                                   & 7.380                                  \\
\textbf{Cycle-GAN \cite{zhu2017unpaired}}                    & 81.564                                  & 87.434                                   & 14.681                             & 7.143                             \\
\textbf{Ours}                         & \textbf{94.333}                         & \textbf{96.081}                          & \textbf{20.086}                    & \textbf{1.316}                    \\ \cline{2-5} 
\textbf{}                             & \multicolumn{4}{c|}{\textbf{DIBCO 2018}}                                                                                                                    \\ \cline{2-5} 
\textbf{Rank 1 \cite{xiong2018historical}}                       & 88.340                                  & 90.240                                   & \textbf{19.110}                    & 4.920                            \\
\textbf{Otsu \cite{otsu1979threshold}}                         & 51.450                                  & 53.050                                   & 9.740                              & 59.070                            \\
\textbf{Niblack \cite{niblack1985introduction}}                      & 41.180                                  & 41.390                                   & 6.790                              & 99.460                            \\
\textbf{Sauvola \cite{sauvola2000adaptive}}                      & 67.810                                  & 74.080                                   & 13.780                             & 17.690                            \\
\textbf{Xiong \cite{xiong2021enhanced}}                        & 88.340                                  & 90.370                                   & 19.110                             & 4.930                             \\
\textbf{Bhunia \cite{bhunia2019improving}}                       & 59.254                                  & 59.178                                   & 11.797                             & 9.555                             \\
\textbf{Suh \cite{suh2020two}}                          & 84.950                                  & 91.577                                   & 17.040                             & 16.861                            \\
\textbf{Soibgui \cite{souibgui2020gan}}                      & 73.700                                        & 75.610                                         & 16.990                                   & 10.210                                  \\
\textbf{Cycle-GAN \cite{zhu2017unpaired}}                    & 72.972                                  & 77.391                                   & 13.462                             & 129.277                           \\
\textbf{Ours}                         & \textbf{89.751}                         & \textbf{93.141}                          & 17.439                             & \textbf{3.824}                    \\ \cline{2-5} 
\textbf{}                             & \multicolumn{4}{c|}{\textbf{DIBCO 2019}}                                                                                                                    \\ \cline{2-5} 
\textbf{Rank 1 \cite{bera2021non}}                       & 72.875                                  & 72.150                                   & 14.475                             & 16.235                            \\
\textbf{Otsu \cite{otsu1979threshold}}                         & 52.800                                  & 52.550                                   & 12.640                             & 24.210                            \\
\textbf{Niblack \cite{niblack1985introduction}}                      & 51.510                                  & 53.860                                   & 10.540                             & 31.050                            \\
\textbf{Sauvola \cite{sauvola2000adaptive}}                      & 42.520                                  & 39.760                                   & 7.710                              & 120.120                           \\
\textbf{Bhunia \cite{bhunia2019improving}}                       & 53.340                                  & 55.995                                   & 11.779                             & 9.256                             \\
\textbf{Suh \cite{suh2020two}}                          & 62.893                                  & 62.726                                   & \textbf{15.584}                             & \textbf{3.362}                             \\
\textbf{Soibgui \cite{souibgui2020gan}}                      & 70.330                                        & 71.470                                         & 12.220                                   & 8.910                                  \\
\textbf{Cycle-GAN \cite{zhu2017unpaired}}                    & 74.916                                  & \textbf{75.189}                          & 14.307                             & 6.814                             \\
\textbf{Ours}                         & \textbf{75.130}                         & 75.101                                   & 14.802                    & 5.248\\
\hline
\end{tabular}
\end{table}

\section{Ablation Studies}
Although our experiments show that the proposed method achieves competitive performance against existing state-of-the-art methods, we would like experimentally prove the benefits brought by the pivotal points of our method - generation of synthetic data while training and the modified Pix2Pix architecture suggested in Section 3.2 (B). Here, we perform our ablation study to verify the advantage of individual modules in the proposed model.

\subsection{W/O Variational Augmentation}

We train our binarization network, Bi-Net without Stage 1 and compare it with the proposed method in Table \ref{tab:abl_1}. We observe that there is a significant drop in performance of the model when we do not incorporate variational augmentation. Given the limited amount of training data, the model fails to learn the various degradations that might be present in real-world historical document images hence it justifies our case for simulating the degradations using a neural network. 

\begin{table}[]
\caption{\label{tab:abl_1}Ablation experiments w/o variational augmentation.}
\setlength{\tabcolsep}{3pt}
\centering
\begin{tabular}{|l|c|c|c|c|}
\hline
\multicolumn{1}{|c|}{\textbf{Method}} & \multicolumn{4}{c|}{\textbf{Evaluation Metrics}}                                                                                                            \\ \hline
\textbf{}                             & \multicolumn{1}{l|}{\textbf{F-Measure $\uparrow$}} & \multicolumn{1}{l|}{\textbf{pF-Measure $\uparrow$}} & \multicolumn{1}{l|}{\textbf{PSNR $\uparrow$}} & \multicolumn{1}{l|}{\textbf{DRD $\downarrow$}} \\ \cline{2-5} 
                                      & \multicolumn{4}{c|}{\textbf{DIBCO 2014}}                                                                                                                    \\ \cline{2-5} 
\textbf{w/o Aug-Net}                  & 91.190                                  & 92.076                                   & 19.758                             & 3.192                             \\
\textbf{Proposed}                     & \textbf{96.860}                         & \textbf{97.450}                          & \textbf{22.250}                    & \textbf{1.530}                    \\ \cline{2-5} 
\textbf{}                             & \multicolumn{4}{c|}{\textbf{DIBCO 2016}}                                                                                                                    \\ \cline{2-5} 
\textbf{w/o Aug-Net}                  & 87.268                                  & 90.844                                   & 17.348                             & 4.142                             \\
\textbf{Proposed}                     & \textbf{94.333}                         & \textbf{96.081}                          & \textbf{20.086}                    & \textbf{1.316}                    \\ \cline{2-5} 
\textbf{}                             & \multicolumn{4}{c|}{\textbf{DIBCO 2018}}                                                                                                                    \\ \cline{2-5} 
\textbf{w/o Aug-Net}                  & 73.968                                  & 77.170                                   & 14.988                             & 118.359                            \\
\textbf{Proposed}                     & \textbf{89.751}                         & \textbf{93.141}                          & \textbf{17.439}                    & \textbf{3.824}                    \\ \cline{2-5} 
\textbf{}                             & \multicolumn{4}{c|}{\textbf{DIBCO 2019}}                                                                                                                    \\ \cline{2-5} 
\textbf{w/o Aug-Net}                  & 64.809                                  & 64.802                                   & 14.458                             & 6.594                             \\
\textbf{Proposed}                     & \textbf{75.130}                         & \textbf{75.101}                          & \textbf{14.802}                    & \textbf{5.248} \\
\hline
\end{tabular}
\end{table}

\subsection{W/O Residual Blocks + PixelShuffle}
We also compare Bi-Net with its parent network Pix2Pix and compare their performances in Table \ref{tab:abl_2}. We see that our proposed architecture outperforms Pix2Pix on DIBCO datasets by a significant margin. Bi-Net's PixelShuffle upsampling preserves line low details better than conventional strided convolutions which is evident by the boost in F-Measure.

\begin{table}[]
\caption{\label{tab:abl_2}Ablation experiments w/o changes in architecture of Bi-Net}
\setlength{\tabcolsep}{4pt}
\centering
\begin{tabular}{|l|c|c|c|c|}
\hline
\multicolumn{1}{|c|}{\textbf{Method}} & \multicolumn{4}{c|}{\textbf{Evaluation Metrics}}                                                                                                            \\ \hline
\textbf{}                             & \multicolumn{1}{l|}{\textbf{F-Measure $\uparrow$}} & \multicolumn{1}{l|}{\textbf{pF-Measure $\uparrow$}} & \multicolumn{1}{l|}{\textbf{PSNR $\uparrow$}} & \multicolumn{1}{l|}{\textbf{DRD $\downarrow$}} \\ \cline{2-5} 
                                      & \multicolumn{4}{c|}{\textbf{DIBCO 2014}}                                                                                                                    \\ \cline{2-5}
\textbf{Pix2Pix}                      & 73.928                                  & 74.606                                   & 18.548                             & 9.059                             \\
\textbf{Proposed}                     & \textbf{91.190}                         & \textbf{92.076}                          & \textbf{19.758}                    & \textbf{3.192}                    \\ \cline{2-5} 
\textbf{}                             & \multicolumn{4}{c|}{\textbf{DIBCO 2016}}                                                                                                                    \\ \cline{2-5} 
\textbf{Pix2Pix}                      & 72.636                                  & 73.689                                   & 13.641                             & 8.800                             \\
\textbf{Proposed}                     & \textbf{87.268}                         & \textbf{90.844}                          & \textbf{17.348}                    & \textbf{4.142}                    \\ \cline{2-5} 
\textbf{}                             & \multicolumn{4}{c|}{\textbf{DIBCO 2018}}                                                                                                                    \\ \cline{2-5} 
\textbf{Pix2Pix}                      & 68.135                                  & 68.516                                   & 9.324                              & \textbf{55.791}                            \\
\textbf{Proposed}                     & \textbf{73.968}                         & \textbf{77.170}                          & \textbf{14.988}                    & 118.359                   \\ \cline{2-5} 
\textbf{}                             & \multicolumn{4}{c|}{\textbf{DIBCO 2019}}                                                                                                                    \\ \cline{2-5} 
\textbf{Pix2Pix}                      & \textbf{68.626}                                  & \textbf{68.735}                                   & 9.478                              & 17.714                            \\
\textbf{Proposed}                     & 64.809                         & 64.802                          & \textbf{14.458}                    & \textbf{6.594}  \\
\hline
\end{tabular}
\end{table}

\section{Conclusion}
In this paper, we have proposed a novel document binarization algorithm that successfully deals binarization of historical document images. Our algorithm couples the strengths of variational inference and paired image-to-image translation and hence is able to perform well in scenarios where training data is scarce. We demonstrate the efficacy of our methods by performing a quantitative analysis of the binarization performance using metrics like F-Measure, pseudo-F-Measure, PSNR and DRD. The experimental results show that the proposed method outperforms traditional and state-of-the-art methods on multiple metrics. Furthermore, we have conducted ablation experiments to prove the merits of our methodology. We observe, however, that our method fails to binarize images where the foreground text has faded away. Furthermore, since we do not analyze the textual information within the image we are unable to restore images where a part of the text is missing or areas where the ink has bloated, rendering the text unreadable. To fix these caveats, future work in this domain focusing on the integration of language processing algorithms within the binarization framework is necessary.


  \bibliographystyle{unsrtnat}
  \bibliography{variational}

%
%
%

\end{document}